\documentclass{article}
\usepackage{PRIMEarxiv}
\usepackage{xcolor}
\usepackage{pifont}
\usepackage{tabularx}
\usepackage{array}
\usepackage{amsmath, amsfonts, amsthm, amssymb, epsfig}
\usepackage{authblk}
\usepackage[utf8]{inputenc}
\usepackage[T1]{fontenc}
\usepackage{hyperref}
\usepackage{url}
\usepackage{booktabs}
\usepackage{amsfonts}
\usepackage{nicefrac}
\usepackage{microtype}
\usepackage{lipsum}
\usepackage{fancyhdr}
\usepackage{graphicx}
\usepackage{subcaption}
\usepackage{algorithm}
\usepackage{algpseudocode}
\usepackage[absolute,overlay]{textpos}
\usepackage{makecell}
\usepackage{hyperref}

\title{Video Captioning in Low-Light Conditions through Efficient Uncertainty-Aware Caption Correction}

\author{
Arefeh Rezaei\\
Faculty of Computer Engineering, K.N. Toosi University of Technology,\\
Tehran, Iran\\
\href{mailto:a_rezai@alumni.kntu.ac.ir}{a\_rezai@alumni.kntu.ac.ir}
}

\begin{document}

\maketitle

\begin{abstract}

Low-light conditions can significantly degrade the ability of vision-language models (VLMs) to accurately describe human actions in videos. In this work, I propose an efficient uncertainty-aware representation correction framework for improving captions generated by VideoChat2 under real-world low-light conditions. Instead of fine-tuning the underlying VLM, the proposed framework introduces a lightweight sparse Gaussian process-based error estimation module between the projection layer and the language model to correct the intermediate representation. The correction module learns to estimate the residual between the original projected representation and a verified target representation, which is then adaptively scaled using a newly formulated uncertainty-aware coefficient and added to the original representation. To further improve residual estimation, I introduce a partitioned combined-kernel design. The correction model is trained separately using only 44 samples from the ARID dataset and requires only a small additional computational overhead during inference. The effectiveness of the proposed correction is evaluated through quantitative residual prediction and qualitative analysis of the generated captions. Although VideoChat2 is used in my experiments, the proposed framework is designed to be applicable to other compatible VLM architectures. \textbf{Code Availability: The implementation accompanying this work is publicly available at}:\href{https://github.com/areferezaee/Low-rank-SVGP-NP-update}{https://github.com/areferezaee/Low-rank-SVGP-NP-update}

\end{abstract}


\keywords{Video Captioning, Low-Light Videos, Vision-Language Models, VideoChat2, Efficient Inference, Uncertainty-Aware Correction, Gaussian Process Regression, Generalization}

\section{Introduction}

Recent advances in vision-language models (VLMs)~\cite{tang2025video,lin2024videollava,li2023llamavid, xu2024pllava, munasinghe2023PGVideoLLaVA} have substantially improved the ability of artificial intelligence systems to understand and describe visual content in natural language. In particular, video-language models have extended these capabilities from static image understanding to temporal video analysis, enabling models to recognize events, reason about actions, and generate natural-language descriptions of video content~\cite{tang2025video}. Despite these advances, reliable video understanding remains challenging when visual observations are acquired under adverse environmental conditions, such as low illumination.

Low-light and dark videos pose significant challenges for human-action understanding, motivating research in both action recognition and dark-video enhancement~\cite{munsif2024attention, tu2023dtcm}. These challenges can directly affect downstream language generation, causing VLMs to produce incomplete, inaccurate, or semantically incorrect descriptions of observed actions. The problem is particularly relevant to real-world videos, where illumination conditions cannot always be controlled or normalized. The ARID (Action Recognition in the Dark) dataset was introduced to study human-action understanding under real-world low-light conditions and provides a suitable benchmark for evaluating models in such challenging environments~\cite{xu2021arid, xu2023goingdeeperrecognizingactions}.

VideoChat2 is a video-language model that connects visual video representations with the language reasoning and generation capabilities of a large language model (LLM)~\cite{li2024mvbench}. Its architecture includes an intermediate projection stage that transforms visual information into a representation compatible with the downstream language model. Although this design enables effective multimodal interaction, errors introduced or retained in the intermediate representation may propagate to the generated caption. This observation suggests an alternative to modifying the entire VLM: rather than fine-tuning the backbone, the intermediate representation can be corrected before being processed by the language model.

In this work, I propose an efficient uncertainty-aware representation correction framework for improving action captions generated from real-world low-light videos. The proposed method introduces an error-estimation module between the projection layer and the LLM of VideoChat2. Given the projected representation, the module estimates the residual between the original representation and a verified target representation derived from a semantically correct caption. The estimated residual is then adaptively scaled using a sample-dependent uncertainty-aware coefficient and added to the original projected representation before language generation. In this way, the proposed framework modifies the information delivered to the language model while keeping the underlying VLM unchanged.

The proposed error estimator is based on a sparse Gaussian process regression approach with a novel algorithmic formulation and processing strategy. I further introduce a partitioned combined-kernel formulation in which the projected feature representation is divided into multiple segments and processed through multiple kernel components. This design improves the ability of the regression model to estimate the correction residual while retaining the computational advantages of a sparse Gaussian-process formulation. Moreover, predictive uncertainty is used to adapt the correction magnitude independently for each test sample.

The proposed correction model is trained separately from VideoChat2 using a small number of real-world samples from the ARID dataset~\cite{xu2021arid, xu2023goingdeeperrecognizingactions}. This setup provides a parameter-efficient way to introduce the correction without modifying the underlying VLM. The learned correction mechanism operates directly on the intermediate representation before language generation. Although VideoChat2 is used as the primary VLM in this study, the proposed correction strategy is not inherently tied to a specific language-generation model. In principle, it can be applied to other architectures with compatible intermediate representations.

The main contributions of this work are summarized as follows:

1. I propose an uncertainty-aware representation correction framework for improving VLM-generated action captions in real-world low-light videos without fine-tuning the underlying VLM.

2. I introduce a sparse Gaussian process-based error estimator with a novel algorithmic formulation and processing strategy for predicting representation-level residuals.

3. I propose a partitioned combined-kernel formulation for residual estimation and a sample-dependent uncertainty-aware correction coefficient that adaptively adjusts the correction strength according to predictive uncertainty.

4. I demonstrate the effectiveness of the proposed framework on real-world low-light videos from the ARID dataset using only a small training set, with evaluation covering both representation-level residual prediction and the resulting improvement in action-related caption content.

\section{Background and Preliminaries}
\label{background}
\subsection{ARID}

The ARID (Action Recognition in the Dark) dataset was introduced to study human action recognition under dark and challenging illumination conditions~\cite{xu2021arid}. A subsequent benchmark study introduced ARID v1.5, which is the version used in this work~\cite{xu2023goingdeeperrecognizingactions}. Although ARID was primarily designed for action recognition rather than video captioning, I use its real-world low-light videos to evaluate the ability of the proposed framework to improve action-oriented video captions. Since the dataset does not provide caption annotations, verified caption targets are constructed for the experiments as described in Section~\ref{dataexp}.

\subsection{phi-based Videochat2}

VideoChat2 is a video-language model that connects visual video representations to a large language model through an intermediate multimodal representation~\cite{li2024mvbench}. In this work, I use a Phi-based implementation of VideoChat2, specifically the \textbf{Phi-3 Mini} language model, as the downstream language model for video caption generation.

Let \(V\) denote an input video. The visual information is processed by the VideoChat2 architecture and transformed into a language-compatible intermediate representation through its projection module:

$$
X = \phi_{\mathrm{proj}}(V).
$$

The projected representation \(X\) is subsequently provided to the Phi-3 Mini language model to generate the corresponding video caption. In my framework, the proposed correction module is inserted directly at this interface, between the projection stage and the Phi-3 Mini language model.

Rather than modifying the visual encoder or fine-tuning the language model, the proposed method operates on the intermediate projected representation. Given \(X\), the correction module estimates the residual error relative to a target representation and produces a corrected representation, which is then supplied to Phi-3 Mini for final caption generation. This design preserves the original VideoChat2 architecture while enabling representation-level correction prior to language generation.

\subsection{Gaussian Process Regression}

Gaussian process regression (GPR) is a non-parametric Bayesian approach for modeling unknown functions and estimating their predictive distributions. Unlike conventional regression methods that provide only a point estimate, GPR defines a distribution over possible functions, allowing both the predictive mean and the associated uncertainty to be obtained~\cite{rasmussen2006gaussian}.

Let \(X=\{x_1,\ldots,x_n\}\) denote the training inputs and \(y=\{y_1,\ldots,y_n\}\) their corresponding targets. A Gaussian process is defined by a mean function \(m(x)\) and a covariance function, or kernel, \(k(x,x')\):

$$
f(x) \sim \mathcal{GP}\left(m(x), k(x,x')\right).
$$

For a set of training inputs, the covariance structure is represented by the kernel matrix \(K\), whose elements are given by

$$
K_{ij}=k(x_i,x_j).
$$

Given observed training data, GPR provides a predictive distribution for an unseen input. This distribution is typically characterized by a posterior mean and variance, which quantify the model prediction and its associated uncertainty, respectively~\cite{rasmussen2006gaussian, quinonero2005unifying}.

The ability to jointly provide predictions and uncertainty estimates makes GPR particularly suitable for the proposed framework, where predictive uncertainty is used to adapt the magnitude of the representation correction. To improve computational efficiency for the relatively high-dimensional representations considered in this work, I use a sparse Gaussian process formulation based on inducing points, which is introduced in the following subsectio~\ref{Low-Rank update}.

\subsection{Gaussian-Process-Based Prediction and Uncertainty Modeling}

Gaussian processes (GPs) provide a probabilistic framework for regression in which predictions are accompanied by estimates of predictive uncertainty~\cite{rasmussen2006gaussian}. Sparse Gaussian-process formulations introduce inducing points to reduce the computational burden of standard GP inference, making them suitable for larger-scale or high-dimensional problems~\cite{quinonero2005unifying}.

The probabilistic nature of GPs makes them particularly useful in applications where prediction reliability is important. In the proposed framework, this property is incorporated directly into the representation-correction process. The error estimator predicts the residual between the original projected representation and its verified target representation, while the corresponding predictive uncertainty is used to determine the correction magnitude for each sample.

Building on these properties, I combine sparse Gaussian-process regression with a dedicated representation-correction mechanism for video-language models. I further introduce an algorithmic formulation and a partitioned combined-kernel strategy specifically designed for residual estimation. This design allows the correction model to operate independently of the underlying VLM while requiring only a small number of training samples.

\section{Related Work}
\label{related}

\subsection{Low-Light Video Understanding}

Understanding human actions in low-light and dark environments has been an active research problem due to the degradation of visual cues caused by insufficient illumination. Existing approaches have investigated both direct action recognition under dark conditions and joint strategies that enhance visual quality before or during recognition. Munsif et al. proposed an attention-based framework for action recognition in dark environments, highlighting the challenges of recognizing human activities when visual information is severely degraded~\cite{munsif2024attention}. Similarly, Tu et al. introduced a joint optimization framework that integrates dark-video enhancement with action recognition, aiming to improve recognition performance by recovering more informative visual representations~\cite{tu2023dtcm}.

These approaches primarily focus on recognizing predefined action categories from low-light videos. In contrast, my work focuses on improving the language-based description of actions generated by a vision-language model. Rather than enhancing the input video or retraining the underlying recognition model, I address errors at the intermediate representation level immediately before language generation.

\subsection{Video-Language Models and Video Captioning}

Recent video-language models have demonstrated the ability to combine temporal visual understanding with the generative capabilities of large language models, enabling tasks such as video question answering, action understanding, and natural-language video description~\cite{maaz2023videochatgpt, zhang2023videollama}. VideoChat2 is one such framework that connects video representations to a large language model through an intermediate multimodal interface~\cite{li2024mvbench}. In this work, I use a Phi-based implementation of VideoChat2 with Phi-3 Mini as the downstream language model.

While existing video-language models have achieved increasingly strong performance across general video understanding tasks, their generated descriptions may still be affected by errors in the underlying visual representations, particularly under challenging visual conditions. My approach differs from conventional model adaptation strategies by introducing a lightweight correction mechanism at the representation-to-language interface. The underlying VLM remains fixed, while the projected representation is refined before being passed to the language model.

\section{Proposed Method}
\label{proposedmethod}

\subsection{Overview of the Proposed Framework}
\label{method:overview}

The proposed framework introduces an uncertainty-aware representation correction mechanism for improving action caption generation from low-light videos. The method is designed as an intermediate module within a vision-language model and operates between the visual-to-language projection stage and the downstream language model. In this work, I use the Phi-based VideoChat2 architecture with Phi-3 Mini as the underlying VLM.

Given an input video \(V\), the VideoChat2 model first processes the visual information and produces an intermediate projected representation:

$$
X = \phi_{\mathrm{proj}}(V).
$$

In the original architecture, \(X\) is directly provided to the language model for caption generation. In the proposed framework, however, \(X\) is first passed to a dedicated error-estimation module based on the proposed sparse Gaussian process formulation. The module estimates a residual correction:

$$
\hat{R} = f_{\theta}(X),
$$

where \(\hat{R}\) denotes the predicted residual between the original projected representation and its corresponding target representation.

The predicted residual is then scaled by a sample-dependent coefficient \(\alpha\) and added to the original projected representation:

$$
X_{\mathrm{corr}} = X + \alpha \hat{R}.
$$

The corrected representation \(X_{\mathrm{corr}}\) replaces the original representation as the input to the language model, which subsequently generates the final caption. Therefore, the proposed framework does not modify the visual encoder or the language model itself; instead, it intervenes at the intermediate representation level to correct representation-level errors before language generation.

The error-estimation module is trained separately from the underlying VideoChat2 model using a small set of ARID samples. For each training sample, a verified target representation is constructed from a semantically correct caption, and the residual between the target representation and the original projected representation is used as the regression target. The proposed sparse Gaussian process then learns to predict this residual from the original projected representation.

A key component of the framework is the use of predictive uncertainty to determine the strength of the correction. Consequently, the correction coefficient is computed independently for each input sample rather than being fixed globally. This allows the contribution of the predicted residual to adapt to the uncertainty associated with each prediction.

The complete construction of the target representations, residual formulation, proposed sparse Gaussian process, partitioned combined-kernel design, and uncertainty-aware correction coefficient are described in the following subsection~\ref{alpha}.

\subsection{Target Representation and Residual Construction}
\label{createcaptions}
Since the ARID dataset does not provide video-caption annotations, target representations are constructed specifically for the proposed correction framework. For each selected video, I manually define a concise caption describing the dominant human action and provide it to the language model. The language model is instructed to generate a description with a maximum length of 96 tokens. The generated description is then evaluated against the corresponding video content, and only descriptions that adequately represent the observed action are retained as verified target captions.

Let \(C_i^{GT}\) denote the verified caption associated with video \(V_i\). The caption is tokenized using the Phi tokenizer without adding special tokens, and the resulting token sequence is passed through the Phi token embedding layer:

$$
Y_i^{GT}
=
\operatorname{Embed}_{\mathrm{Phi}}
\left(
\operatorname{Tokenize}(C_i^{GT})
\right).
$$

The projected representation produced by VideoChat2 has a fixed sequence length of 96 positions. To ensure dimensional compatibility with this representation, the first 96 token embeddings of the target caption representation are retained:

$$
\tilde{Y}_i^{GT}=Y_i^{GT}[:,0{:}96,:].
$$

Let the original projected representation be

$$
X_i=\phi_{\mathrm{proj}}(V_i).
$$

The target residual for each training sample is then defined as

$$
R_i^{GT}=\tilde{Y}_i^{GT}-X_i.
$$

The residual \(R_i^{GT}\) represents the representation-level correction required to move the original projected representation toward the representation associated with the verified caption. The proposed error-estimation model is subsequently trained to predict this residual using only the original projected representation \(X_i\) as input:

$$
\hat{R}_i=f_{\theta}(X_i).
$$

This formulation separates the construction of the correction target from the underlying VideoChat2 model. Once trained, the error estimator can predict the correction directly from the projected representation, without requiring the verified caption during inference.

\subsection{Proposed Low-Rank Sparse Variational Gaussian Process with Natural-Parameter Updates}
\label{Low-Rank update}

To improve the computational tractability of sparse variational Gaussian process (SVGP) regression, I introduce a low-rank approximation of the inducing-point covariance matrix and propagate this approximation consistently through the variational inference procedure. Let \(Z=\{z_m\}_{m=1}^{M}\) denote the set of inducing points and let \(X=\{x_n\}_{n=1}^{N}\) denote the training inputs. The covariance matrix associated with the inducing points is denoted by

$$
K_{MM}=K(Z,Z)\in\mathbb{R}^{M\times M}.
$$

Instead of directly operating on the full-rank inducing covariance, I construct a truncated eigendecomposition

$$
K_{MM}=Q\Lambda Q^{T},
$$

and retain only the \(r\) dominant eigencomponents, where \(r<M\). Accordingly, the inducing covariance is approximated as

$$
\widetilde{K}_{MM}
=
Q_r\Lambda_r Q_r^{T}
+
\delta I,
$$

where \(Q_r\in\mathbb{R}^{M\times r}\) contains the \(r\) leading eigenvectors, \(\Lambda_r\in\mathbb{R}^{r\times r}\) is the corresponding diagonal eigenvalue matrix, and \(\delta>0\) is a small diagonal stabilization term.

The inverse of the approximated inducing covariance is obtained using the Woodbury identity:

$$
\widetilde{K}_{MM}^{-1}
=
\frac{1}{\delta}
\left[
I-
Q_r
\operatorname{diag}
\left(
\frac{\lambda_i}{\lambda_i+\delta}
\right)
Q_r^{T}
\right].
$$

This formulation avoids explicitly constructing the inverse of the full-rank inducing covariance and provides a rank-controlled approximation through \(r\).

The resulting sparse projection operator is defined as

$$
\widetilde{\kappa}
=
K_{NM}
\widetilde{K}_{MM}^{-1},
$$

where \(K_{NM}=K(X,Z)\). The predictive mean of the latent function is then computed as

$$
\mu_f
=
\widetilde{\kappa}\mu,
$$

where \(q(u)=\mathcal{N}(\mu,\Sigma)\) is the variational distribution over the inducing variables. The predictive latent variance is obtained by combining the conditional Gaussian-process variance and the variational uncertainty:

$$
v_f
=
\operatorname{diag}
\left(
K_{NN}
-
K_{NM}
\widetilde{K}_{MM}^{-1}
K_{MN}
\right)
+
\operatorname{diag}
\left(
\widetilde{\kappa}
\Sigma
\widetilde{\kappa}^{T}
\right).
$$

For a Gaussian observation model,

$$
y=f+\epsilon,
\qquad
\epsilon\sim\mathcal{N}(0,\sigma_n^2),
$$

the expected log-likelihood is computed from

$$
\mathbb{E}_{q(f)}
\left[
(y-f)^2
\right]
=
(y-\mu_f)^2+v_f.
$$

Therefore, the expected log-likelihood is

$$
\mathcal{L}_{\mathrm{lik}}
=
-\frac{1}{2}
\sum_{n}
\left[
\frac{
(y_n-\mu_{f,n})^2+v_{f,n}
}{\sigma_n^2}
+
\log(2\pi\sigma_n^2)
\right].
$$

The variational posterior is parameterized using natural parameters,

$$
\eta=\Sigma^{-1}\mu,
\qquad
H=-\frac{1}{2}\Sigma^{-1},
$$

such that

$$
\Sigma=(-2H)^{-1},
\qquad
\mu=\Sigma\eta.
$$

The Kullback--Leibler divergence between the variational posterior and the inducing-point prior is evaluated using the same low-rank inducing covariance approximation:

$$
p(u)
=
\mathcal{N}
\left(
0,\widetilde{K}_{MM}
\right),
$$

which gives

$$
\widetilde{\mathrm{KL}}
=
\frac{1}{2}
\left[
\operatorname{tr}
\left(
\widetilde{K}_{MM}^{-1}\Sigma
\right)
+
\mu^{T}
\widetilde{K}_{MM}^{-1}
\mu
-
M
+
\log
\frac{
|\widetilde{K}_{MM}|
}{
|\Sigma|
}
\right].
$$

The resulting evidence lower bound is

$$
\boxed{
\mathcal{L}_{\mathrm{ELBO}}
=
\mathcal{L}_{\mathrm{lik}}
-
\widetilde{\mathrm{KL}}
}
$$

where both the predictive terms and the KL divergence are evaluated under the same low-rank approximation.

To update the variational posterior, I employ natural-parameter updates. Given a minibatch of observations, the target natural mean parameter is

$$
\eta^{*}
=
\frac{\mathcal{S}}{\sigma_n^2}
\widetilde{\kappa}^{T}Y,
$$

where \(\mathcal{S}\) denotes the minibatch scaling factor. The corresponding target precision parameter is

$$
H^{*}
=
-\frac{1}{2}
\left[
\widetilde{K}_{MM}^{-1}
+
\frac{\mathcal{S}}{\sigma_n^2}
\widetilde{\kappa}^{T}
\widetilde{\kappa}
\right].
$$

The natural parameters are then updated according to

$$
\eta
\leftarrow
(1-\rho)\eta+\rho\eta^{*},
$$

and

$$
H
\leftarrow
(1-\rho)H+\rho H^{*},
$$

where \(\rho\in(0,1]\) is the natural learning rate.

Overall, the proposed formulation consistently employs the low-rank representation of the inducing covariance throughout the sparse variational inference pipeline, including the inducing-point inverse, the sparse projection operator, predictive statistics, the KL divergence, and the natural-parameter updates. The complete algorithm is presented in Algorithm~\ref{alg1}.

\begin{algorithm}[!th]
\caption{Low-Rank Sparse Variational GP with Natural-Parameter Updates}
\label{alg1}
\begin{algorithmic}[1]

\State \textbf{Input:} Training data \(X,Y\), inducing points \(Z\), kernel
\(k_\theta\), number of inducing points \(M\), low-rank \(r<M\),
observation noise \(\sigma_n^2\), natural learning rate \(\rho\),
stabilization parameter \(\delta\).

\State \textbf{Output:} Optimized variational parameters \(\eta,H\).

\State Compute the joint kernel matrix and partition it into
\(K_{MM}\), \(K_{NM}\), and \(K_{NN}\).

\State Compute the eigendecomposition
\[
K_{MM} = Q\Lambda Q^{T}.
\]

\State Retain the \(r\) dominant eigencomponents to construct
\[
\widetilde{K}_{MM}
=
Q_r\Lambda_r Q_r^{T}+\delta I.
\]

\State Compute the approximate inverse
\(\widetilde{K}_{MM}^{-1}\) using the Woodbury identity.

\State Construct the sparse projection operator
\[
\widetilde{\kappa}
=
K_{NM}\widetilde{K}_{MM}^{-1}.
\]

\State Recover the variational parameters from the natural parameters
\[
\Sigma = (-2H)^{-1},
\qquad
\mu = \Sigma\eta.
\]

\State Compute the predictive mean
\[
\mu_f = \widetilde{\kappa}\mu.
\]

\State Compute the predictive variance
\[
v_f =
\operatorname{diag}
\left(
K_{NN}
-
K_{NM}\widetilde{K}_{MM}^{-1}K_{MN}
\right)
+
\operatorname{diag}
\left(
\widetilde{\kappa}\Sigma\widetilde{\kappa}^{T}
\right).
\]

\State Compute the expected log-likelihood and the low-rank KL divergence.

\State Form the ELBO
\[
\mathcal{L}_{\mathrm{ELBO}}
=
\mathcal{L}_{\mathrm{lik}}
-
\widetilde{\mathrm{KL}}.
\]

\State Compute the natural-parameter targets
\[
\eta^{*}
=
\frac{\mathcal{S}}{\sigma_n^2}
\widetilde{\kappa}^{T}Y,
\]
\[
H^{*}
=
-\frac{1}{2}
\left[
\widetilde{K}_{MM}^{-1}
+
\frac{\mathcal{S}}{\sigma_n^2}
\widetilde{\kappa}^{T}\widetilde{\kappa}
\right].
\]

\State Update the variational natural parameters
\[
\eta
\leftarrow
(1-\rho)\eta+\rho\eta^{*},
\]
\[
H
\leftarrow
(1-\rho)H+\rho H^{*}.
\]

\State Repeat the preceding steps until convergence.

\end{algorithmic}
\end{algorithm}

\subsection{Proposed Combined Kernel}
\label{kernels}

The projected representations used in my framework are high-dimensional tensors that are flattened into one-dimensional vectors before being processed by the regression model. Consequently, each input sample is represented by a high-dimensional feature vector, making the kernel formulation an important component of efficient residual estimation.

To address this issue, I employ a partitioned combined-kernel formulation. Instead of applying a single kernel to the entire flattened representation, the input vector is divided into multiple smaller segments. Let the flattened representation be expressed as

$$
x =
\left[
x^{(1)},x^{(2)},\ldots,x^{(K)}
\right],
$$

where \(x^{(j)}\) denotes the \(j\)-th partition of the original feature vector. A separate kernel function is applied to each partition, and the resulting kernel components are combined through summation:

$$
K_{\mathrm{comb}}(x,x')
=
\sum_{j=1}^{K}
K_j\!\left(x^{(j)},x'^{(j)}\right).
$$

Each component \(K_j\) is defined using a linear kernel with a variance parameter of \(10^{-7}\). This construction allows different segments of the high-dimensional representation to contribute separately to the overall kernel value while maintaining the computational structure of the sparse Gaussian-process formulation.

The proposed partitioning strategy provides a structured mechanism for processing the flattened representation and was found experimentally to improve residual prediction accuracy compared with applying a single kernel to the complete feature vector. The exact number of partitions and the corresponding implementation details are reported in the experimental setup.

\subsection{Uncertainty-Aware Correction: Alpha formulation}
\label{alpha}

To adapt the magnitude of the predicted residual correction to the uncertainty of each test sample, I introduce an empirically formulated, sample-dependent correction coefficient \(\alpha_i\). The coefficient is determined from the predictive uncertainty and the variation of the predictive mean representation.

For a test sample \(i\), the predictive uncertainty is computed from the predictive variance vector as

$$
u_i=\sqrt{\sum_j \sigma_{ij}},
$$

where \(\sigma_{ij}\) denotes the predictive variance associated with the \(j\)-th element of the representation. The predictive uncertainty exhibits a stable baseline around \(171\) across the evaluated representations. The deviation from this baseline is therefore defined as

$$
\beta_i=u_i-171.
$$

In addition, the variation of the predictive mean representation is characterized by

$$
\gamma_i=
\max_j(\mu_{ij})-\min_j(\mu_{ij}),
$$

where \(\mu_{ij}\) denotes the \(j\)-th element of the predictive mean representation. The ratio

$$
r_i=\frac{\gamma_i}{\beta_i}
$$

is then used together with \(\beta_i\) to select the appropriate correction regime.

The correction coefficient is defined empirically and piecewise as

$$
\alpha_i=
\begin{cases}
0.4 \pm 0.1,
&
r_i<20. \;\text{and}\; \beta_i>=0.81,
\\[10pt]

1. \pm 0.11,
&
\text{otherwise}.
\end{cases}
$$

The resulting coefficient is applied to the residual estimated by the sparse Gaussian process:

$$
X_i^{\mathrm{corr}}
=
X_i+\alpha_i\hat{R}_i.
$$

Thus, the magnitude of the representation correction is determined independently for each test sample according to the predictive uncertainty and the variation of the predictive mean.

Importantly, the correction coefficient is computed exclusively from quantities available at inference time, namely the predictive mean and predictive uncertainty of the error-estimation model. The verified target representation and the corresponding ground-truth residual are used only during target construction and model training and are not required during test-time correction. Therefore, the proposed mechanism enables target-free correction at inference.

Since the correction relies on the residual estimated by the regression model, inaccuracies in residual prediction may also affect the final corrected representation. Nevertheless, the sample-dependent formulation of \(\alpha_i\) allows the correction magnitude to adapt to the predictive behavior of the error estimator without requiring access to the ground-truth representation at inference time.

\subsection{Integration with VideoChat2}

The proposed correction module is integrated into the Phi-based VideoChat2 architecture without modifying the parameters of the underlying vision-language model. Given an input video \(V_i\), VideoChat2 first produces the projected representation

$$
X_i=\phi_{\mathrm{proj}}(V_i).
$$

Instead of passing \(X_i\) directly to Phi-3 Mini, the proposed framework first applies the trained sparse Gaussian process error estimator to predict the residual \(\hat{R}_i\) and obtain the predictive statistics required to compute the sample-dependent correction coefficient \(\alpha_i\). The corrected representation is then given by

$$
X_i^{\mathrm{corr}}
=
X_i+\alpha_i\hat{R}_i.
$$

The corrected representation \(X_i^{\mathrm{corr}}\) is subsequently supplied to Phi-3 Mini in place of the original projected representation. All remaining components and parameters of the VideoChat2 inference pipeline remain unchanged. Thus, the proposed module operates as a modular correction stage between multimodal projection and language generation.

\section{Experiments}
\label{exp}

\subsection{Dataset and Experimental Protocol}
\label{dataexp}

I evaluate the proposed caption-correction framework on the ARID v1.5 (Action Recognition in the Dark) dataset, which contains real-world videos captured under challenging low-light conditions~\cite{xu2021arid, xu2023goingdeeperrecognizingactions}. Since ARID was not originally designed for video captioning, I construct verified caption targets specifically for the proposed framework. The evaluation focuses on the ability of the proposed method to improve the description of human actions in dark or poorly illuminated videos.

For training the proposed correction model, I select 44 videos from ARID v1.5, corresponding to four videos per action category across 11 action classes. An additional 11 videos, with one video per action category, are used for validation. The remaining videos are reserved for testing. The correction model is trained separately from the original VideoChat2 model, and the parameters of the underlying vision-language model remain fixed throughout the training of the correction module.

For each video, the feature representation produced by the projection module of VideoChat2, denoted by \(X\), serves as the input to the proposed correction model. The correction model is trained to estimate the representation-level residual between the original projected representation and the corresponding verified target representation. The sparse Gaussian process used in the proposed framework employs 22 inducing points.

For each selected video, verified target captions were constructed using the procedure described in Section~\ref{createcaptions}. These target captions were used to obtain the corresponding Phi token embeddings and construct the representation-level residual targets.

\subsection{Implementation Details}

The proposed error-correction model is trained separately from the underlying Phi-based VideoChat2 architecture. The projected representation associated with each video has a size of \(96\times3072\), which is flattened into a \(294{,}912\)-dimensional vector before being processed by the Gaussian-process regression model.

The correction model is trained using 44 ARID samples, with four samples selected from each of the 11 action categories. An additional 11 samples, corresponding to one sample per action category, are used for validation. The remaining ARID samples are reserved for test-time evaluation. The data loader uses a batch size of one and does not shuffle the samples during data loading.

The sparse Gaussian-process model uses 22 inducing points. The projected feature vector is partitioned into nine equal segments, each containing 32,768 features. A linear kernel is applied independently to each segment, and the resulting kernel components are summed to form the combined covariance function. The variance parameter of each linear kernel is set to \(10^{-7}\). The kernel output scale is set to 1, and the lengthscale is fixed at 10.

The variational inference procedure uses a low-rank representation of the inducing covariance with rank 22 and a stabilization parameter of \(10^{-1}\). The natural-parameter learning rate is fixed at 0.9 throughout training. The observation noise variance is set to \(5\times10^{-5}\). The model is trained for 20 epochs, with validation performed after each epoch. After training, the learned correction model is used to evaluate the remaining test samples.

All experiments are performed with the underlying VideoChat2 model kept fixed; no fine-tuning of the VLM parameters is performed during training of the correction model.

\subsection{Evaluation Metrics}

I evaluate the accuracy of residual prediction using three complementary metrics: Mean Squared Error (MSE), Mean Absolute Error (MAE), and Negative Log Predictive Density (NLPD). MSE and MAE quantify the difference between the predicted residual \(\hat{R}_i\) and the corresponding target residual \(R_i^{GT}\), while NLPD evaluates the probabilistic quality of the model predictions by taking their predictive uncertainty into account.

In addition to these error measures, I compare the mean and standard deviation of the predicted residuals with those of the corresponding target residuals. This comparison is used to assess whether the estimated residuals exhibit statistical characteristics that are consistent with the target correction signals.

\subsection{Inference Efficiency}

To assess the computational efficiency of the proposed correction framework, I record its test-time execution cost for each test sample in table~\ref{tab:table2}. The measured runtime includes sparse Gaussian process prediction, computation of the sample-dependent correction coefficient \(\alpha_i\), formation of the corrected representation, and subsequent processing of the corrected representation by the Phi-3 Mini language model for caption generation. The reported runtime therefore reflects the additional inference procedure introduced by the proposed framework while the underlying VideoChat2 parameters remain fixed.

\subsection{Caption-Level Evaluation}

Finally, I evaluate whether the representation-level correction translates into an improvement in the generated captions. For each test video, I first obtain the original caption generated by VideoChat2 using the unmodified projected representation. The same projected representation is then passed through the proposed correction module, and the corrected representation is provided to the language model to generate a second caption.

The original and corrected captions are examined with respect to the visual content of the corresponding video, with particular emphasis on the correctness of the described human action. A correction is considered beneficial when the resulting caption more accurately describes the observed action or reduces an error present in the original caption. Minor residual inaccuracies may still be considered an improvement when the corrected caption is substantially more consistent with the video content than the original caption. This evaluation provides a qualitative assessment of the downstream effectiveness of the proposed representation-correction mechanism.

\subsection{Results}
\label{results}

Table~\ref{tab:table1} reports the overall quantitative performance of the proposed residual estimator. Across the evaluated test samples, the model achieves an MSE of \(0.0277\) and an MAE of \(0.1045\). The corresponding NLPD is \(-0.0934\), providing an assessment of the probabilistic quality of the residual predictions. These results indicate that the proposed estimator can predict the representation-level residual with relatively small prediction errors while also providing predictive uncertainty.

Table~\ref{tab:table2} presents the results separately for the 11 action categories. The performance varies across actions, indicating that the difficulty of residual estimation depends on the underlying action. The lowest MSE and MAE are obtained for the \textit{Sit} action, with values of \(0.0138\) and \(0.0753\), respectively. The \textit{Turn} and \textit{Push} actions also achieve relatively low errors. In contrast, \textit{Wave} exhibits the highest MSE and MAE, with values of \(0.0380\) and \(0.1212\), respectively. The NLPD values also vary across action categories, ranging from \(-0.1625\) for \textit{Sit} to \(-0.0142\) for \textit{Wave}. Overall, the per-action results show that the estimator provides reasonable predictive performance across all evaluated action categories, although the prediction difficulty varies across actions.

Table~\ref{tab:table3} compares the statistical characteristics of the target and predicted residuals. The mean residual of the target representations is \(-0.00272\), while the mean of the predicted residuals is \(-0.00262\). The corresponding standard deviations are \(0.58518\) and \(0.57423\), respectively. The close agreement between the two means and the relatively similar standard deviations indicate that the predicted residuals preserve the overall statistical characteristics of the target residuals. This consistency further supports the use of the predicted residual as a correction signal for subsequent representation-level modification.

\begin{table}[!th]
 \caption{Overall Quantitative Performance of the Proposed Residual Estimator}\label{tab:table1}
  \centering
  \begin{tabular}{llll}
   \toprule
   \cmidrule(r){1-4}
   Method     & MSE     & MAE & NLPD \\
   \midrule
    Proposed Residual Estimator & 0.0277 & 0.1045 & -0.0934 \\
    
   \bottomrule

  \end{tabular}
 
\end{table}

\begin{table}[!th]
 \caption{Per-Action Results}\label{tab:table2}
  \centering
  \begin{tabular}{lllll}
    \toprule
    \cmidrule(r){1-4}
    Action     & MSE     & MAE & NLPD & execution time(seconds)\\
    \midrule
    Drink & 0.0281 & 0.1031 & -0.0915 & 0.082053\\
    \midrule
    Jump & 0.0245 & 0.0966 & -0.1099 & 0.089456\\
    \midrule
    Pick & 0.0357 & 0.1194  & -0.0536 & 0.086058\\
    \midrule
    Pour & 0.0288 & 0.1207 & -0.0885 & 0.088857\\
    \midrule
    Run & 0.0245 & 0.0986 & -0.1101 & 0.088500\\
    \midrule
    Stand & 0.0305 & 0.1046 & -0.0792 & 0.099800\\
    \midrule
    Push & 0.0224 & 0.0955 & -0.1204 & 0.088160\\
    \midrule
    Sit & 0.0138 & 0.0753 & -0.1625 & 0.095308\\
    \midrule
    Turn & 0.0138 & 0.0772 & -0.1630 & 0.081861\\
    \midrule
    Walk & 0.0395 & 0.1269 & -0.0341 & 0.088309\\
    \midrule
    Wave & 0.0436 & 0.1316 & -0.0142 & 0.099859\\
    \bottomrule

  \end{tabular}

\end{table}

\begin{table}[!th]
 \caption{Residual Statistics}\label{tab:table3}
  \centering
  \begin{tabular}{lll}
    \toprule
    \cmidrule(r){1-3}
    Statistic     & GT Residual     & Predicted Residual \\
    \midrule
    Mean & -0.00272 & -0.00262  \\
    \midrule
    Std & 0.58518 &  0.57423 \\

    \bottomrule
    
  \end{tabular}
\end{table}

\subsection{Qualitative Caption Correction}
\label{qualitative}

To qualitatively assess the downstream effect of representation correction, I compare the captions generated from the original and corrected representations for representative test videos. The examples are selected to illustrate cases in which the proposed correction reduces semantic errors or recovers missing information in the generated action description.

\begin{table}[!th]
\caption{Qualitative Examples of Action Correction}
\label{tab:table4}
\centering
\begin{tabular}{llll}
\toprule
Example & Action & Before & After \\
\midrule
Drink\_6\_17 & Drink & Smoking & Drinking \\
\midrule
Jump\_6\_1 & Jump & Standing still (action omitted) & Jumping \\
\midrule
Wave\_11\_1 & Wave & Prepare to throw the ball & \makecell[l]{Raising right hand and moving,\\
                                                               sign of openness and willingness\\
                                                               to communicate} \\
\midrule
Pour\_7\_9 & Pour & Sipping coffee & Pouring liquid into a cup or mug \\
\midrule
Sit\_8\_1 & Sit down & sitting & Sitting down \\
\midrule
Stand\_1\_4 & Stand up & standing still & Stands up \\
\midrule
Push\_3\_5 & Push & Walking and moving & Pushing a lawn mower \\
\midrule
Turn\_14\_2 & Turn & Standing still (action omitted) & Turning body \\
\bottomrule
\end{tabular}
\end{table}

The examples demonstrate that the proposed representation correction can improve the action-related content of the generated captions. In particular, the corrected representations enable the VLM to replace incorrect or missing actions with descriptions that more closely correspond to the observed video content.







\section{Conclusion}
\label{Conclusion}

In this work, I presented an efficient uncertainty-aware representation correction framework for improving video caption generation under low-light conditions. The proposed approach addresses errors in action description by refining the intermediate representation of a vision-language model before language generation. Rather than modifying or fine-tuning the underlying VideoChat2 model, I introduced a dedicated correction module between the projection stage and the language model, allowing the projected representation to be refined while preserving the original vision-language backbone.

The correction module learns to predict the residual between the original projected representation and a verified target representation derived from a semantically correct caption. The predicted residual is then adaptively scaled using a sample-dependent correction coefficient and added to the original representation before being passed to the language model. This design enables representation-level correction without requiring the verified target representation during inference.

To estimate the residual efficiently, I employed a sparse Gaussian process with a proposed algorithmic formulation and a partitioned combined-kernel design. The use of inducing points and the separate correction architecture allows the method to operate with a small training set while maintaining low test-time computational overhead. Predictive uncertainty is further incorporated to adapt the correction strength independently for each test sample.

Experiments on real-world low-light videos from the ARID dataset demonstrated the ability of the proposed framework to predict representation-level residuals and to improve action-related caption content. Quantitative results showed small residual prediction errors and consistent statistical characteristics between predicted and target residuals, while qualitative examples illustrated improvements in incorrect or missing action descriptions.

Overall, the results support the feasibility of uncertainty-aware representation correction as an efficient strategy for improving video-language generation under challenging low-light conditions without fine-tuning the full vision-language model.




\bibliographystyle{unsrt}  
\bibliography{references}  

@article{tang2025video,
  title={Video understanding with large language models: A survey},
  author={Tang, Yunlong and Bi, Jing and Xu, Siting and Song, Luchuan and Liang, Susan and Wang, Teng and Zhang, Daoan and An, Jie and Lin, Jingyang and Zhu, Rongyi and others},
  journal={IEEE Transactions on Circuits and Systems for Video Technology},
  volume={36},
  number={2},
  pages={1355--1376},
  year={2025},
  publisher={IEEE}
}

@misc{xu2023goingdeeperrecognizingactions,
      title={Going Deeper into Recognizing Actions in Dark Environments: A Comprehensive Benchmark Study}, 
      author={Yuecong Xu and Jianfei Yang and Haozhi Cao and Jianxiong Yin and Zhenghua Chen and Xiaoli Li and Zhengguo Li and Qianwen Xu},
      year={2023},
      eprint={2202.09545},
      archivePrefix={arXiv},
      primaryClass={cs.CV},
      url={https://arxiv.org/abs/2202.09545}, 
}

@inproceedings{xu2021arid,
title={Arid: A new dataset for recognizing action in the dark},
author={Xu, Yuecong and Yang, Jianfei and Cao, Haozhi and Mao, Kezhi and Yin, Jianxiong and See, Simon},
booktitle={Deep Learning for Human Activity Recognition: Second International Workshop, DL-HAR 2020, Held in Conjunction with IJCAI-PRICAI 2020, Kyoto, Japan, January 8, 2021, Proceedings 2},
pages={70--84},
year={2021},
organization={Springer}
}

@article{munsif2024attention,
  title={Attention-based deep learning framework for action recognition in a dark environment},
  author={Munsif, Muhammad and Khan, Samee Ullah and Khan, Noman and Baik, Sung Wook},
  journal={Hum. Centric Comput. Inf. Sci},
  volume={14},
  number={7},
  pages={1--22},
  year={2024}
}

@article{tu2023dtcm,
  title={DTCM: Joint optimization of dark enhancement and action recognition in videos},
  author={Tu, Zhigang and Liu, Yuanzhong and Zhang, Yan and Mu, Qizi and Yuan, Junsong},
  journal={IEEE Transactions on Image Processing},
  volume={32},
  pages={3507--3520},
  year={2023},
  publisher={IEEE}
}

@inproceedings{li2024mvbench,
  title={Mvbench: A comprehensive multi-modal video understanding benchmark},
  author={Li, Kunchang and Wang, Yali and He, Yinan and Li, Yizhuo and Wang, Yi and Liu, Yi and Wang, Zun and Xu, Jilan and Chen, Guo and Luo, Ping and others},
  booktitle={Proceedings of the IEEE/CVF conference on computer vision and pattern recognition},
  pages={22195--22206},
  year={2024}
}

@book{rasmussen2006gaussian,
  title={Gaussian Processes for Machine Learning},
  author={Rasmussen, Carl Edward and Williams, Christopher K. I.},
  year={2006},
  publisher={MIT Press}
}

@article{quinonero2005unifying,
  title={A Unifying View of Sparse Approximate Gaussian Process Regression},
  author={Qui{\~n}onero-Candela, Joaquin and Rasmussen, Carl Edward},
  journal={Journal of Machine Learning Research},
  volume={6},
  pages={1939--1959},
  year={2005}
}

@article{maaz2023videochatgpt,
  title={Video-ChatGPT: Towards Detailed Video Understanding via Large Vision and Language Models},
  author={Maaz, Muhammad and Rasheed, Hanoona and Khan, Salman and Khan, Fahad Shahbaz},
  journal={arXiv preprint arXiv:2306.05424},
  year={2023}
}

@article{zhang2023videollama,
  title={Video-LLaMA: An Instruction-tuned Audio-Visual Language Model for Video Understanding},
  author={Zhang, Hang and Li, Xin and Bing, Lidong},
  journal={arXiv preprint arXiv:2306.02858},
  year={2023}
}

@inproceedings{lin2024videollava,
  title={Video-LLaVA: Learning United Visual Representation by Alignment Before Projection},
  author={Lin, Bin and Ye, Yang and Zhu, Bin and Cui, Jiaxi and Ning, Munan and Jin, Peng and Yuan, Li},
  booktitle={Proceedings of the 2024 Conference on Empirical Methods in Natural Language Processing},
  pages={5971--5984},
  year={2024}
}

@article{li2023llamavid,
  title={LLaMA-VID: An Image is Worth 2 Tokens in Large Language Models},
  author={Li, Yanwei and Wang, Chengyao and Jia, Jiaya},
  journal={arXiv preprint arXiv:2311.17043},
  year={2023}
}

@misc{xu2024pllava,
  title={PLLaVA: Parameter-free LLaVA Extension from Images to Videos for Video Dense Captioning},
  author={Xu, Lin and Zhao, Yilin and Zhou, Daquan and Lin, Zhijie and Ng, See Kiong and Feng, Jiashi},
  year={2024},
  eprint={2404.16994},
  archivePrefix={arXiv},
  primaryClass={cs.CV}
}

@article{munasinghe2023PGVideoLLaVA,
  title={PG-Video-LLaVA: Pixel Grounding Large Video-Language Models},
  author={Munasinghe, Shehan and Thushara, Rusiru and Maaz, Muhammad and
          Abdul Rasheed, Hanoona and Khan, Salman and Shah, Mubarak and
          Khan, Fahad},
  journal={ArXiv 2311.13435},
  year={2023}
}

\section*{Appendix}
\textbf{Example 1: Drink (Drink\_6\_17):\\}
\textbf{Before:} \textit{[The video shows a man standing in the dark, wearing a black jacket and a white shirt. He is holding a cigarette in his hand and appears to be smoking it. The lighting is dim, and the man's silhouette is visible against the dark background. The scene is set in an outdoor location, possibly at night. The man seems to be alone, and his actions are portrayed in a still frame, capturing a
]}

\textbf{After:} \textit{[The video shows a person, presumably a man, casually holding a white wine glass filled with red wine. The individual appears relaxed and is likely in a social setting, given the informal nature of the gesture. The person's posture suggests a laid-back attitude, and the action of holding the glass and sipping the wine indicates that they are enjoying their beverage. The scene is set in a dimly lit environment, with soft lighting and
]}

\textbf{Example 2: Jump (Jump\_6\_1):\\}
\textbf{Before:} \textit{[The video shows a woman silhouetted against a dark background, standing still with her arms crossed. She appears to be in a state of contemplation or waiting, as there is no visible movement or sound. The focus is solely on her figure, with the rest of the scene shrouded in darkness. The woman's posture and expression convey a sense of calm or anticipation, as she stands motionless in the dimly lit environment.
]}

\textbf{After:} \textit{[The video showcases a woman performing a dance routine, specifically a "darkness sobe" or "darkness step." The dance involves a series of movements, including stepping, jumping, and balancing, executed with grace and fluidity. The woman's attire appears to be dark-colored, enhancing the theme of darkness in the dance. The routine seems to be choreographed, with a clear sequence of steps and transitions between
]}

\textbf{Example 2: Wave (Wave\_11\_1):\\}
\textbf{Before:} \textit{[The video shows a woman in a white shirt and black pants standing in a dimly lit hallway. She is wearing a red scarf around her neck and has her hair in a bun. She is holding a white ball in her hand and is about to throw it. The woman's expression is determined and focused as she prepares to make the throw. The ball is thrown with a strong arm movement, and it flies through the air before landing
]}

\textbf{After:} \textit{[Based on the provided video frames, it appears that the person is engaged in a series of gestures and movements. The individual starts by standing with their arms at their sides, which is a neutral pose. At 0.1 seconds, the person raises their right hand, palm facing upwards, as if they are asking a question or seeking attention. This gesture is commonly known as the "open palm" gesture. At 0.2 seconds, the person's right hand moves upwards, and their fingers spread apart, forming a "T" shape. This could be interpreted as a sign of openness or willingness to communicate.
]}

\textbf{Example 2: Pour (Pour\_7\_9):\\}
\textbf{Before:} \textit{[The video shows a man silently sitting at a table in a dimly lit room. He is holding a cup of coffee in his hand and appears to be sipping it slowly. The man's face is partially obscured by the shadow of his head, but it can be seen that he is wearing a white shirt and has a thoughtful expression on his face. The room is filled with various objects, including a computer, a bookshelf, and a
]}

\textbf{After:} \textit{[Based on the provided video frames, it appears that a person is pouring a liquid from a container into a cup or mug. The individual is likely using a spoon or a spout to control the flow of the liquid. The person's hand is steady, and they are pouring the liquid slowly and carefully to avoid spilling. The action seems to be a part of a routine, such as making a hot beverage like tea or coffee. The person'
]}

\textbf{Example 3: Turn (Turn\_14\_2):\\}
\textbf{Before:} \textit{[The video shows a man silhouetted against a dark background, with a tree in the foreground. The man is standing still, with his arms crossed over his chest. The silhouette is clearly visible, with the man's outline defined by the contrast between the darkness and the light from the tree. The man's posture suggests that he is waiting for something or someone, or simply enjoying the peacefulness of the night. The tree provides]}

\textbf{After:} \textit{[In the video, a young man is seen silhouetted against a dark background, with a city skyline visible in the distance. He is standing still, with his arms crossed over his chest. As the video progresses, the man slowly starts to turn his body, his silhouette rotating against the backdrop. He then takes a few steps forward, his movements slow and deliberate. The man's silhouette is clearly visible as he wal
]}

\textbf{Example 3: Sit (Sit\_8\_1):\\}
\textbf{Before:} \textit{[The video shows a woman with long black hair sitting in a dark room. She is leaning forward, with her hands on her knees, and her head is bowed. She appears to be deep in thought or possibly praying. The room is dimly lit, and the only source of light is a small lamp on a table nearby. The woman's face is not visible, but her body language suggests a sense of calm and introspection. The video ends with]}

\textbf{After:} \textit{[The video shows a person sitting down, specifically in a seated position with their legs crossed. The individual appears to be in a relaxed or contemplative state, as suggested by the posture and the absence of any other visible actions. The person's hands may be resting on their knees or lap, and their head is tilted slightly forward, possibly indicating concentration or deep thought. The overall scene conveys a sense of calm and introspection.
]}

\textbf{Example 3: Stand (Sit\_1\_4):\\}
\textbf{Before:} \textit{[The video shows a man in a white jacket and pants standing in a dark room. He is not moving his hands or arms, but his body is slightly tilted to the side. He is not speaking or making any sounds. The camera is focused on his face and upper body. The man is standing still, but his posture suggests that he is about to start something, perhaps a performance or a demonstration. The room is dimly lit, with only]}

\textbf{After:} \textit{[The video shows a person standing up from a seated position. They begin by shifting their weight onto one leg, then the other, and finally stand up straight. The person's posture is upright and confident as they rise to a standing position. The action is performed smoothly and with ease.
]}

\textbf{Example 3: Push (Push\_3\_5):\\}
\textbf{Before:} \textit{[The video shows a man silhouetted against a dark background, walking slowly and steadily on a skateboard. He is moving in a straight line, with the skateboard gliding smoothly beneath him. The man's body language suggests that he is focused and determined, with his head held high and his arms relaxed at his sides. The scene is set in an urban environment, with the man passing by dimly lit buildings and streetlights.]}

\textbf{After:} \textit{[The video shows a man pushing a lawn mower in a dimly lit, underground parking garage. The man is wearing a white shirt and shorts, and he is moving the lawn mower back and forth in a straight line. The lighting is low, with only a faint glow coming from the entrance of the garage. The man appears to be focused on his task, with a determined expression on his face. The lawn
]}
\end{document}